\documentclass{article}
\usepackage{spconf,amsmath,graphicx}
\usepackage{multirow}
\usepackage{bm}
\usepackage{mathrsfs}
\usepackage[]{algorithm2e}
\usepackage{caption}
\title{Controllable Time-Delay Transformer for Real-Time Punctuation Prediction and Disfluency Detection}
\name{Qian Chen, Mengzhe Chen, Bo Li, Wen Wang}
\address{DAMO Academy, Alibaba Group \\
\{tanqing.cq,~~mengzhe.cmz,~~shiji.lb,~~w.wang\}@alibaba-inc.com}

\begin{document}
\ninept
\maketitle
\begin{abstract}
With the increased applications of automatic speech recognition (ASR) in recent years, it is essential to automatically insert punctuation marks and remove disfluencies in transcripts, to improve the readability of the transcripts as well as the performance of subsequent applications, such as machine translation, dialogue systems, and so forth. In this paper, we propose a Controllable Time-delay Transformer (CT-Transformer) model that jointly completes the punctuation prediction and disfluency detection tasks in real time. The CT-Transformer model facilitates freezing partial outputs with controllable time delay to fulfill the real-time constraints in partial decoding required by subsequent applications. We further propose a fast decoding strategy to minimize latency while maintaining competitive performance. Experimental results on the IWSLT2011 benchmark dataset and an in-house Chinese annotated dataset demonstrate that the proposed approach outperforms the previous state-of-the-art models on F-scores and achieves a competitive inference speed.

\end{abstract}

\section{Introduction}
\label{sec:intro}
Spoken language transcripts generated by automatic speech recognition (ASR) systems usually have no punctuation marks. 
And for spontaneous speech, ASR transcripts often include a lot of speech disfluencies. However, many subsequent applications, such as machine translation and dialogue systems, are usually trained on well-formed text with proper punctuation marks and without disfluencies. 
Hence, there is a significant mismatch between the training corpora and the actual speech transcript input for these applications, causing dramatic performance degradation.
In addition, the lack of punctuation marks and the presence of disfluencies reduce the readability of speech transcripts.
Consequently, predicting punctuation and detecting disfluencies (and removing detected disfluencies) have become crucial post-processing tasks for speech transcripts.

One example of speech transcript is ``I want a flight to Boston um to Denver''. 
For punctuation prediction, we annotate whether there is a specific type of punctuation mark after a word, such as period, comma, etc. In this case, there is a period after the word ``Denver''.
The annotation of disfluency includes the reparandum and interregnum. Reparandum includes words that are corrected by the following words or are to be discarded. Reparandum includes repetition, repair, and restart. Interregnum includes filled pauses, discourse markers, etc. In this case, the phrase ``to Boston'' is annotated as reparandum and ``um'' is annotated as interregnum.

A critical challenge for punctuation prediction and disfluency detection for real-time spoken language processing systems is latency. 
For example, simultaneous translation systems \cite{DBLP:conf/acl/MaHXZLZZHLLWW19} require fixed partial post-processed speech transcript and decode it partially (prefix-to-prefix framework) to minimize latency.
In this work, we tackle the challenge of reducing the latency from two aspects, the modeling approach and the decoding strategy, while achieving a high accuracy performance. Previous state-of-the-art approaches use a Transformer encoder-decoder model for punctuation prediction~\cite{DBLP:conf/icassp/YiT19} and disfluency detection~\cite{DBLP:conf/aaai/DongWYCXX19}. The encoder-decoder model consists of the encoder and the auto-regressive decoder, which prevents the model from massive parallelization during inference. Hence, it is difficult to employ such a model in a real-time punctuation prediction and disfluency detection system due to its low inference efficiency. 
Past research \cite{DBLP:conf/emnlp/WangSN14,DBLP:conf/interspeech/BaronSS02} showed that jointly modeling punctuation prediction and disfluency detection can improve their generalization capabilities, enhance the overall efficiency of the pipeline, and avoid error propagation. 
Inspired by the success of self-attention mechanisms for sequence labeling tasks \cite{DBLP:conf/aaai/TanWXCS18}, we propose a Controllable Time-delay Transformer (CT-Transformer) model which jointly models punctuation prediction and disfluency detection. In order to achieve punctuation prediction and disfluency detection in real time, the proposed CT-Transformer model only uses the encoder part of the Transformer encoder-decoder model structure~\cite{DBLP:conf/icassp/YiT19,DBLP:conf/aaai/DongWYCXX19}. Longer context is preferred for better prediction and detection performance, but it also results in higher latency. Compared to cutting off context during inference, CT-Transformer provides a principled way for freezing partial outputs with controllable time delay to fulfill the real-time constraints in partial decoding required by subsequent applications.

Previous work studied different decoding strategies to reduce the latency for real-time spoken language processing systems, including overlapping windows \cite{DBLP:conf/iwslt/ChoNW12}, streaming input scheme \cite{cho2015punctuation}, and overlapped-chunk split and merging strategy \cite{DBLP:journals/corr/abs-1908-02404}.
However, the input text for inference in these decoding strategies does not always begin with the first word of a sentence. Hence these strategies may ignore crucial context information for predicting punctuation and detecting disfluency. 
Tilk et al. \cite{DBLP:conf/interspeech/TilkA16} proposed a decoding strategy which always begins with the first word of a sentence. 
Since this approach partitions the input sequence into 200-word slices, it cannot be used in real-time streaming systems due to its high latency.
We propose a fast decoding strategy to minimize latency while maintaining competitive performance. 
This strategy guarantees that the input text for inference always begins with the first word of a sentence. 
Meanwhile, to reduce the computational complexity, the strategy dynamically throws away a history that is too long based on already predicted punctuation marks.

In addition, most previous approaches to punctuation prediction and disfluency detection are supervised approaches and rely heavily on human-annotated speech transcripts, which are expensive to obtain. 
To tackle the training data bottleneck, we investigate transfer learning to exploit existing large-scale well-formed text corpora.

Our contributions can be summarized as follows: 
1) We propose a Controllable Time-delay Transformer (CT-Transformer) model to jointly model punctuation prediction and disfluency detection through multi-task learning. CT-Transformer provides a principled approach that facilitates freezing partial outputs with controllable time delay to fulfill the real-time constraints in partial decoding required by subsequent applications. To the best of our knowledge, this is the first work that employs self-attention networks for jointly modeling punctuation prediction and disfluency detection, and is the first work that provides the controllable time-delay capability for these tasks.
2) We propose a fast decoding strategy to minimize latency while maintaining competitive performance for stream processing.
3) We investigate transfer learning to utilize existing large-scale well-formed text corpora.
4) Experimental results on the IWSLT2011 benchmark test set and an in-house Chinese annotated dataset show that our approach outperforms the previous state-of-the-art models on F-scores with a competitive latency, and fulfill the real-time constraints.

\vspace{-0.1in}
\section{Related Work}
Punctuation prediction models can be categorized into three major categories, including hidden inter-word event detection~\cite{DBLP:journals/taslp/LiuSSHOH06} (n-gram language models \cite{DBLP:conf/icassp/BeefermanBL98}, Hidden Markov Models (HMMs) \cite{christensen2001punctuation}), sequence labeling by assigning a punctuation mark to each word \cite{DBLP:conf/interspeech/UeffingBV13,DBLP:conf/interspeech/ZelaskoSMSCD18} (conditional random fields (CRFs) \cite{DBLP:conf/emnlp/LuN10}, convolutional neural networks (CNNs) \cite{DBLP:conf/lrec/CheWYM16}, recurrent neural networks (RNNs) \cite{DBLP:conf/interspeech/TilkA15} and its variants \cite{DBLP:conf/interspeech/TilkA16,DBLP:conf/interspeech/YiTWL17}), and sequence-to-sequence modeling in which the source is unpunctuated text and the target is punctuated text \cite{DBLP:conf/iwslt/PeitzFMN11} or sequences of punctuation marks \cite{DBLP:conf/icassp/YiT19,DBLP:conf/slt/Klejch0R16}.

For disfluency detection, previous methods can be categorized into four categories: sequence labeling, parsing-based, noisy channel model, and encoder-decoder model.
The sequence labeling method labels each word as fluent or not using different model structures, including CRFs \cite{DBLP:conf/interspeech/OstendorfH13}, HMMs \cite{DBLP:journals/taslp/LiuSSHOH06}, RNNs \cite{DBLP:conf/interspeech/HoughS15} or others \cite{DBLP:conf/naacl/Georgila09,DBLP:journals/corr/abs-1908-05378}. 
Noisy channel models use the similarity between reparandum and repair as an indicator of disfluency \cite{DBLP:conf/acl/JohnsonC04,DBLP:conf/acl/LouJ17}.
The parsing-based approaches jointly model syntactic parsing and disfluency detection tasks \cite{DBLP:conf/emnlp/RasooliT13,DBLP:conf/emnlp/YoshikawaSM16}.
The encoder-decoder models defines disfluency detection as a sequence-to-sequence problem \cite{DBLP:journals/csl/NeubigAMK12,DBLP:conf/aaai/DongWYCXX19}.

Previous work also investigated masked self-attention mechanisms for natural language processing. 
Shen et al. \cite{DBLP:conf/aaai/ShenZLJPZ18} proposed diag-disabled mask, forward mask, and backward mask for language understanding. 
Song et al. \cite{DBLP:journals/corr/abs-1811-00253} investigated local mask and directional mark in Transformer for machine translation. 
However, these masked self-attention mechanisms are different from our proposed controllable time-delay self-attention, as explained in Section~\ref{subsec:ct-transformer}.

\vspace{-0.1in}
\section{Proposed Approach}
\label{sec:approach}
\begin{figure}[t]
\centering
\includegraphics[width=0.40\textwidth]{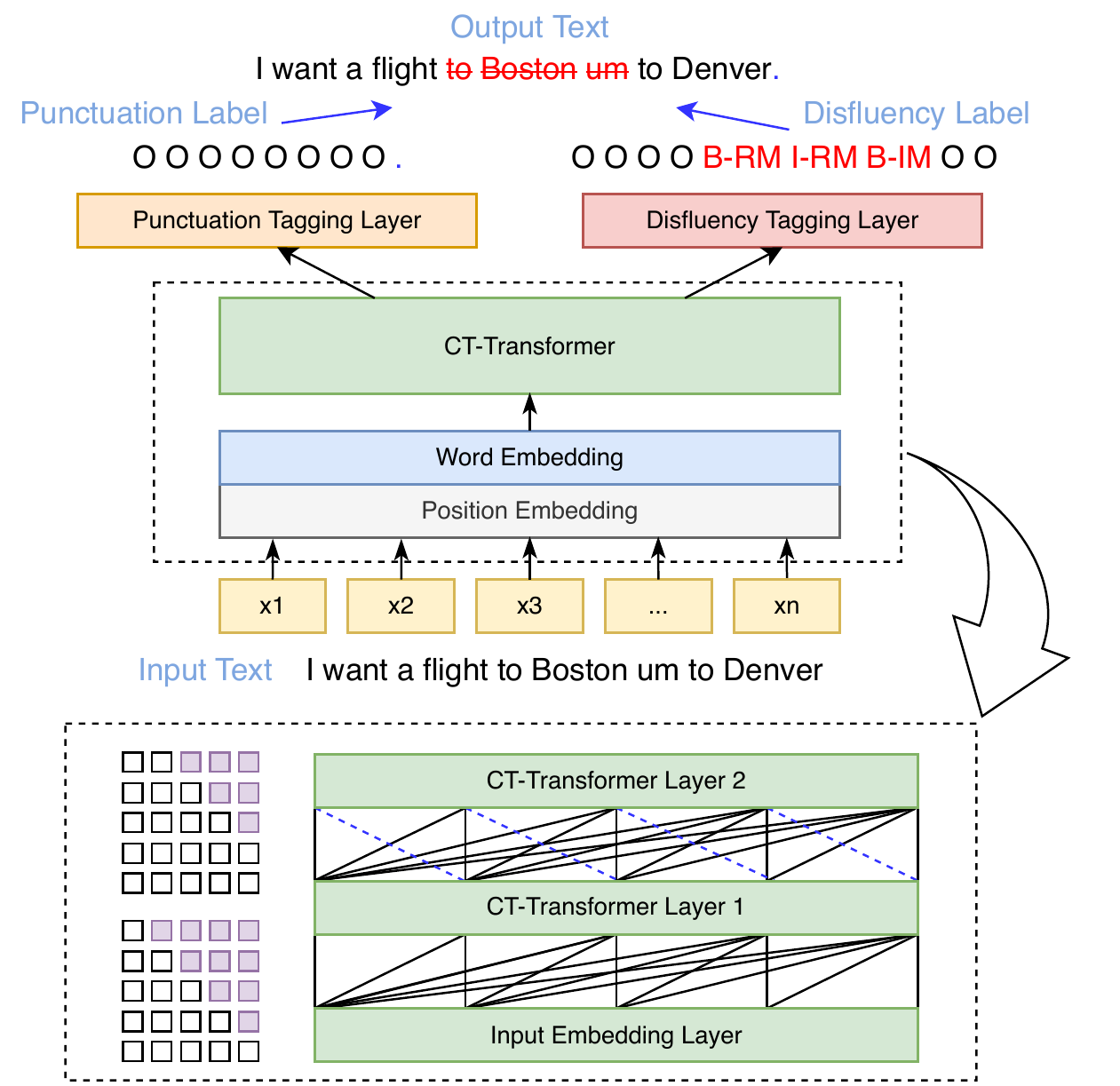}
\caption{The architecture of the CT-Transformer model for real-time joint punctuation prediction and disfluency detection.}
\label{fig:model}
\end{figure}

The proposed model is illustrated in Figure~\ref{fig:model}. 
The inputs are transcripts, e.g., ``I want a flight to Boston um to Denver''. 
The outputs are punctuation and disfluency labels using the BIO scheme \cite{DBLP:conf/acl-vlc/RamshawM95}, e.g., ``O O O O O O O O .'' and ``O O O O B-RM I-RM B-IM O O'', where ``B'', ``I'', ``O'' denote Beginning, Inside, and Outside of a text segment, and ``RM'' and ``IM''  denote reparandum and interregnum.

\vspace{-0.1in}
\subsection{Model Architecture}
\label{subsec:architecture}
The input embedding consists of word embeddings and position embeddings (sinusoidal position encoding).
The encoder consists of a stack of $N$ layers. 
Each layer has two sub-layers, i.e., the multi-head self-attention sub-layer and the fully connected feed-forward network sub-layer.
The output layers consist of the punctuation tagging layer and the disfluency tagging layer. 
The encoder is shared between the two tasks while the tagging layers are separate for each task, following the multi-task learning paradigm.
The final hidden states of the encoder are fed into the corresponding softmax layers for classifying over the punctuation labels and disfluency labels, respectively. The total loss is the summation of the cross entropy losses of punctuation prediction and disfluency detection.
\vspace{-0.1in}
\subsection{Controllable Time-delay Self-attention}
\label{subsec:ct-transformer}
Different from the full sequence self-attention Transformer encoder in the original Transformer model \cite{DBLP:conf/nips/VaswaniSPUJGKP17}, we propose a controllable time-delay self-attention mechanism to encourage the model to depend on future words in a shorter time window instead of the full sequence, thus the partial outputs can be fixed to fulfill the real-time constraints in partial decoding required by subsequent applications.
The original self-attention mechanism builds upon the scaled dot-product attention, operating on query $Q$, key $K$, and value $V$:
\begin{align}
\small
\mathrm{Attention}(Q,K,V) = \mathrm{softmax}(\frac{Q K^T}{\sqrt{d_k}})V \,,
\label{equ:attention}
\end{align}
\noindent where $d_k$ is the dimension of the keys. 
To encourage punctuation prediction and disfluency detection to depend on the future words in a shorter time window, we need to block the flow of information from distant future words into the encoder. To achieve this goal, we modify the scaled dot-product attention by masking out (setting to $-\infty$) all values in the input to the softmax layer which correspond to the unwanted distant future words (illegal connections). We denote this new mechanism the controllable time-delay self-attention (denoted CT self-attention).
Equation~\ref{equ:attention} is modified as
\begin{align}
\small
\mathrm{Attention}(Q,K,V) = \mathrm{softmax}(\frac{Q K^T}{\sqrt{d_k}} + M)V \,,
\label{equ:ct-attention}
\end{align}
where matrix $M$ is
\begin{align}
\small
M_{ij} =
\begin{cases}
0& i+L \geq j\\
-\infty & \text{otherwise}
\end{cases}
\label{equ:mask}
\end{align}
In the mask $M$, there is only attention for position j for the fixed length $L$ of future words and all history words with respect to the current position i. This is illustrated in the bottom of Figure~\ref{fig:model}.
The fixed length $L$ is $0$ in the ``CT-Transformer Layer 1'', which is usually used in the encoder-decoder framework to preserve the auto-regressive property \cite{DBLP:conf/nips/VaswaniSPUJGKP17}.
The fixed length $L$ is $1$ in the ``CT-Transformer Layer 2'', thus the total number of the seen future words is $1=0+1$ here.
The maximum number of the seen future words for each word in CT-Transformer is $L_{all} = \sum L_i, i \in [1, \dots, N]$, where $L_i$ denotes the fixed length in Layer $i$ of the encoder.
The CT self-attention is an extension of previous forward mask \cite{DBLP:conf/aaai/ShenZLJPZ18} and local mask \cite{DBLP:journals/corr/abs-1811-00253}. If $L=0$ in the CT self-attention, the CT self-attention degenerates into the forward mask; if there is only attention for the fixed length $L$ of history words in the CT self-attention, it becomes the local mask.
\vspace{-0.1in}
\subsection{Fast Decoding Strategy}
To simulate the actual streaming scenario for real-time punctuation prediction and disfluency detection systems, we remove segmentation from transcripts. Hence there is only a single input utterance in our evaluations, keeping the same setup with previous work \cite{DBLP:conf/lrec/CheWYM16}.
During training, half of the samples (utterances) are appended with randomly truncated segments to encourage the model not to always predict the end-of-utterance punctuation in the end. We propose a fast decoding strategy with a low frame rate, as shown in Algorithm \ref{alg:decoding}, to reduce latency while maintaining a competitive performance.

\begin{algorithm}[ht]
\footnotesize
\renewcommand{\arraystretch}{0.9}
 \KwData{The low frame rate $F$; the number of look-ahead words $T$; input speech transcripts stream $\mathcal{S}$; input buffer $\mathcal{B}$;}
 Empty $\mathcal{B}$ \;
 \While{not at the end of $\mathcal{S}$}{
  Pop $F$ new words from $\mathcal{S}$ and push them to $\mathcal{B}$\;
  Conduct inference on $\mathcal{B}$ by the CT-Transformer model\;
  \If{the number of words after the first predicted end-of-sentence mark (i.e., period or question mark) is equal or larger than $T$}{
   Remove the words preceding the first predicted end-of-sentence mark in $\mathcal{B}$;
   }
 }
\caption{The fast decoding strategy.}
\label{alg:decoding}
\end{algorithm}

\vspace{-0.2in}
\section{Experiments}
\subsection{Datasets}
We evaluate punctuation prediction on the English IWSLT2011 benchmark dataset. 
We evaluate both punctuation prediction and disfluency detection on an in-house Chinese dataset. 
The IWSLT2011 benchmark contains three types of punctuation marks (comma, period, and question mark), following the data organization and using the same tokenized data by Che et al. \cite{DBLP:conf/lrec/CheWYM16}\footnote{https://github.com/IsaacChanghau/neural\_sequence\_labeling}. Since no public Chinese corpus with both punctuation and disfluency annotations is available at the time of this work, we annotate transcripts for about 240K spoken utterances with punctuation and disfluency annotations, and randomly partition the data into the train, dev, and test sets. We use Jieba\footnote{https://github.com/fxsjy/jieba} for word segmentation. The punctuation annotations consist of four types of punctuation marks (comma, period, question mark, and enumeration comma). We use the BIO scheme to annotate the two types of disfluencies, reparandum and interregnum, for sequence labeling of disfluency detection. Note that in this work, the train, dev, and test sets of both IWSLT2011 and Chinese datasets are manual transcripts. In future work, we will evaluate the robustness of the proposed approach on ASR transcripts.

\begin{table}[ht]
\renewcommand{\arraystretch}{0.9}
\begin{center}
\scalebox{0.8}{
\begin{tabular}{l l r r r}
\hline
\multicolumn{1}{l}{\textbf{Dataset}} & 
\multicolumn{1}{l}{\textbf{Split}} & \multicolumn{1}{l}{\textbf{\#Words}} &
\multicolumn{1}{l}{\textbf{\#Punc.}} & \multicolumn{1}{l}{\textbf{\#Disf.}}\\
\hline
\multirow{4}{*}{\textbf{IWSLT2011}} & Train & 2M & 301K & -  \\
& Train-pretrain & 3.6B  & 399M & -\\
 & Dev & 296K & 43K & - \\
& Test & 13K & 2K & -\\
\hline
\multirow{4}{*}{\textbf{Chinese}} & Train & 5M & 745K & 443K  \\
 & Train-pretrain & 1.6B & 253M & 100M\\
 & Dev &  132K & 18K & 12K\\
& Test & 93K & 15K & 9K \\
\hline
\end{tabular}
}
\end{center}
\caption{The statistics of the train, train-pretrain, dev, and test sets for the IWSLT2011 and in-house Chinese datasets.}
\label{tab:stat}
\end{table}

In order to reduce reliance on expensive annotations on speech transcripts, we explore the pre-training and fine-tuning transfer learning method and existing large-scale well-formed text for pre-training for our tasks. We crawl public Internet resources (news, Wikipedia, question-answering data, discussion forums, etc) to create two large-scale corpora for English and Chinese, respectively. 
We use heuristic rules to map all the punctuation marks in the crawled text to the punctuation marks for the English IWSLT2011 dataset and the Chinese dataset, respectively.
For the crawled Chinese text, we randomly insert reparandum and interregnum using heuristic rules similar to \cite{DBLP:journals/corr/abs-1908-05378} for disfluency detection. 
We use Jieba for word segmentation for the crawled Chinese text. 
The processed English and Chinese crawled text are used for pre-training, denoted IWSLT2011 Train-pretrain and Chinese Train-pretrain datasets, respectively.
The data statistics are summarized in Table~\ref{tab:stat}.
We evaluate punctuation and disfluency detection using token-based precision (P), recall (R), F$_1$-score (F$_1$), following previous works~\cite{DBLP:conf/interspeech/TilkA15,DBLP:conf/acl/LouJ17}. 

\begin{table*}[htb]
\renewcommand{\arraystretch}{0.9}
\begin{center}
\scalebox{0.8}{
\begin{tabular}{c c c c c c c c c c c c c}
\hline
\multirow{2}{*}{\textbf{Model}} & 
\multicolumn{3}{c}{\textbf{Comma}} & \multicolumn{3}{c}{\textbf{Period}} &
\multicolumn{3}{c}{\textbf{Question}} & \multicolumn{3}{c}{\textbf{Overall}}\\
& P & R & F$_1$ & P & R & F$_1$ & P & R & F$_1$ & P & R & F$_1$ \\
\hline
T-LSTM \cite{DBLP:conf/interspeech/TilkA15} & 49.6 & 41.4 & 45.1 & 60.2 & 53.4 & 56.6 & 57.1 & 43.5 & 49.4 & 55.0 & 47.2 & 50.8 \\
T-BRNN-pre \cite{DBLP:conf/interspeech/TilkA16} & 65.5 & 47.1 & 54.8 & 73.3 & 72.5 & 72.9 & 70.7 & 63.0 & 66.7 & 70.0 & 59.7 & 64.4 \\
BLSTM-CRF \cite{DBLP:conf/interspeech/YiTWL17} & 58.9 & 59.1 & 59.0 & 68.9 & 72.1 & 70.5 & 71.8 & 60.6 & 65.7 & 66.5 & 63.9 & 65.1 \\
Teacher-Ensemble \cite{DBLP:conf/interspeech/YiTWL17} & 66.2 & 59.9 & 62.9 & 75.1 & 73.7 & 74.4 & 72.3 & 63.8 & 67.8 & 71.2 & 65.8 & 68.4 \\
Self-attention-word-speech \cite{DBLP:conf/icassp/YiT19} & 67.4 & 61.1 & 64.1 & 82.5 & 77.4 & 79.9 & 80.1 & 70.2 & 74.8 & 76.7 & 69.6 & 72.9 \\
\hline 
BLSTM w/o Pretrain & 53.1 & 48.3 & 50.6 & 66.9 & 70.0 & 68.4 & 70.0 & 45.7 & 55.3 & 60.6 & 58.6 & 59.6\\
Full-Transformer w/o Pretrain & 56.8 & 56.0 & 56.4 & 68.5 & 75.6 & 71.9 & 59.6 & 67.4 & 63.3 & 62.8 & 65.7 & 64.2\\
CT-Transformer w/o Pretrain&   53.3 & 61.8 & 57.2 & 76.2 & 64.3 & 69.7 & 67.5 & 58.7 & 62.8 & 62.9 & 62.9 & 62.9\\
BLSTM  & 64.4 & 60.2 & 62.3 & 73.7 & 83.4 & 78.2 & 71.7 & 71.7 & 71.7 & 69.5 & 71.6 & 70.6  \\
Full-Transformer & 68.8 & 67.3 & 68.1 & 73.9 & 85.5 & 79.3 & 66.7 & 78.3 & 72.0 & 71.4 & 76.3 & 73.8 \\
CT-Transformer & 68.8 & 69.8 & 69.3 & 78.4 & 82.1 & 80.2 & 76.0 & 82.6 & 79.2 & 73.7 & 76.0 & 74.9\\
\hline
\end{tabular}
}
\end{center}
\caption{The results of punctuation prediction in terms of P(\%) ,R(\%) , F1(\%) on the IWSLT2011 test set.}
\label{tab:result:en}
\end{table*}

\vspace{-0.1in}
\subsection{Training Details}
For IWSLT2011, we only have punctuation annotations, so there is only the punctuation tagging layer in Figure~\ref{fig:model} for this dataset.
The encoder consists of a stack of 6 layers. There are $h = 8$ parallel attention layers, or heads. For each of these heads, we use $d_k = d_v = d_{model}/h = 64$. 
The dimension of the inner-layer is 2048.
Adam~\cite{DBLP:journals/corr/KingmaB14} with gradient clipping and warm-up is used for optimization. 
The fixed length $L$ in CT-Transformer is set to $9$, and $L_{i}=0, i \in [1,\dots,N-1]$, $L_N=9$. 
The low frame rate is $F=3$ and the number of look-ahead words after end-of-sentence mark is $T=6$. For IWSLT2011, the batch size is 600 for pre-training and 32 for fine-tuning. 
For the Chinese dataset, the batch size is 600 for both pre-training and fine-tuning.
All these hyper-parameters are optimized on the development sets based on accuracy and latency.

\begin{table*}[htb]
\renewcommand{\arraystretch}{0.9}
\begin{center}
\scalebox{0.8}{
\begin{tabular}{c c c c c c c c c c c c c c c c c}
\hline
\multirow{2}{*}{\textbf{Model}} & 
\multicolumn{3}{c}{\textbf{Comma}} & \multicolumn{3}{c}{\textbf{Period}} &
\multicolumn{3}{c}{\textbf{Question}} &
\multicolumn{3}{c}{\textbf{Enum. Comma}} & \multicolumn{3}{c}{\textbf{Overall}} & 
\multirow{2}{*}{\textbf{Inference Time}}\\
& P & R & F$_1$ & P & R & F$_1$ & P & R & F$_1$ & P & R & F$_1$ & P & R & F$_1$\\
\hline
BLSTM  & 58.9 & 43.9 & 50.3 & 59.7 & 58.1 & 58.9 & 77.0 & 58.8 & 66.7 & 59.8 & 16.5 & 25.9 & 60.2 & 48.8 & 53.9 & 1112.9s ($\times1.0$) \\
Full-Transformer & 61.9 & 50.7 & 55.8 & 60.5 & 64.7 & 62.5 & 74.2 & 68.6 & 71.3 & 64.5 & 30.9 & 41.8 & 62.1 & 55.9 & 58.8 & 676.7s ($\times1.6$)\\
CT-Transformer & 60.8 & 53.5 & 56.9 & 63.8 & 59.7 & 61.7 & 76.3 & 63.0 & 69.0 & 63.4 & 25.2 & 36.1 & 62.7 & 55.3 & 58.8 & 585.8s ($\times1.9$) \\
\hline
\end{tabular}
}
\end{center}
\caption{The results of punctuation prediction and the total inference time on the in-house Chinese test set.}
\label{tab:result:cn}
\end{table*}

\begin{table}[htb]
\renewcommand{\arraystretch}{0.9}
\begin{center}
\scalebox{0.75}{
\begin{tabular}{c c c c c c c c c c}
\hline
\multirow{2}{*}{\textbf{Model}} & 
\multicolumn{3}{c}{\textbf{Inter.}} & \multicolumn{3}{c}{\textbf{Repar.}} &
\multicolumn{3}{c}{\textbf{Either}} \\
& P & R & F$_1$ & P & R & F$_1$ & P & R & F$_1$\\
\hline
BLSTM  & 78.7 & 71.5 & 75.0 & 74.9 & 22.5 & 34.6 & 84.1 & 57.0 & 67.9 \\
Full-Transformer & 77.7 & 74.2 & 75.9 & 74.6 & 28.2 & 40.9 & 83.1 & 61.2 & 70.5\\
CT-Transformer  &77.0 & 74.9 & 75.9 & 74.4 & 27.8 & 40.5 & 82.4 & 61.5 & 70.5 \\
\hline
\end{tabular}
}
\end{center}
\caption{The results of disfluency detection on the Chinese test set.}
\label{tab:result:cn:disf}
\end{table}

\vspace{-0.1in}
\subsection{Results and Discussions}
We evaluate the proposed \textbf{CT-Transformer} model, together with two counterparts for punctuation prediction and disfluency detection.
\textbf{BLSTM} denotes the model that replaces the CT-Transformer block in Figure~\ref{fig:model} with bidirectional LSTM, which has a hidden size of 512 and 6 layers, keeping the size comparable with that of CT-Transformer.
\textbf{Full-Transformer} denotes the model that replaces the CT-Transformer block in Figure~\ref{fig:model} with a full sequence Transformer. The results of these models on the IWSLT2011 test set are reported in the last group in Table~\ref{tab:result:en}. ``Overall'' denotes the micro-average of scores for all types of punctuation marks.
Both Full-Transformer and CT-Transformer outperform BLSTM on overall F$_1$.
CT-Transformer achieves better overall F$_1$ than Full-Transformer (74.9\% versus 73.8\%). Removing pre-training degrades the performance of CT-Transformer and Full-Transformer significantly, and CT-transformer without pre-training yields worse F$_1$ than Full-Transformer (62.9\% versus 64.2\%).  The first group of models and results in Table~\ref{tab:result:en} are cited from previous works. T-LSTM \cite{DBLP:conf/interspeech/TilkA15} used a uni-directional LSTM model and T-BRNN-pre \cite{DBLP:conf/interspeech/TilkA16} used a bidirectional RNN with attention. 
BLSTM-CRF and Teacher-Ensemble are the best single and ensemble models in \cite{DBLP:conf/interspeech/YiTWL17}, respectively.
The previous state-of-the-art model is Self-attention-word-speech \cite{DBLP:conf/icassp/YiT19}, which used a full sequence Transformer encoder-decoder model with pre-trained word2vec and speech2vec embedding features. 
Our proposed CT-Transformer significantly outperforms the previous state-of-the-art model (Self-attention-word-speech) (74.9\% versus 72.9\%). 

We also compare CT-Transformer with the two counterparts on the Chinese dataset for both punctuation prediction (Table~\ref{tab:result:cn}) and disfluency detection (Table~\ref{tab:result:cn:disf}).
Firstly, we observe that CT-Transformer with multi-task learning outperforms CT-Transformer trained only on the punctuation prediction task (58.8\% versus 58.4\%). Hence, we only compare models with multi-task learning in the following experiments.
As shown in Table \ref{tab:result:cn}, CT-Transformer achieves a significantly better overall F$_1$ than BLSTM (58.8\% versus 53.9\%) and is comparable with Full-Transformer. 
Using Intel Xeon Platinum 8163 CPU for inference, compared with BLSTM, Full-Transformer is 1.6x faster in the inference time and CT-Transformer is 1.9x faster, as shown in Table~\ref{tab:result:cn}~\footnote{Note that for the first group of models in Table~\ref{tab:result:en}, T-BRNN-pre partitions the input sequence into 200-word slices, which cannot be used in real-time streaming systems due to its high latency. The other works did not report their inference time or release the source code to test the inference time.}.
Compared with the latest overlapped-chunk split and merging strategy \cite{DBLP:journals/corr/abs-1908-02404}, the proposed fast decoding with CT-Transformer has lower latency (10 words versus 20 words) and better overall punctuation prediction F$_1$ (58.8\% versus 57.8\%)~\footnote{The overlapped-chunk split and merging strategy uses a chunk size 30, sliding window 15, and min\_words\_cut 10 as in \cite{DBLP:journals/corr/abs-1908-02404}.}.

Table~\ref{tab:result:cn:disf} shows the results of detecting reparandum, interregnum, and either, on the in-house Chinese test set. 
We observe that it is much easier to detect interregnum than reparandum. 
For detecting either disfluency type, CT-Transformer achieves a significantly better F$_1$ than BLSTM (70.5\% versus 67.9\%), and is comparable with Full-Transformer.
Since the previous state-of-the-art model for disfluency detection is a full-sequence Transformer model~\cite{DBLP:journals/corr/abs-1908-05378}, these results show that CT-Transformer achieves comparable accuracy to the previous state of the art with lower latency. 

\begin{figure}[htb]
\centering
\includegraphics[width=0.45\textwidth]{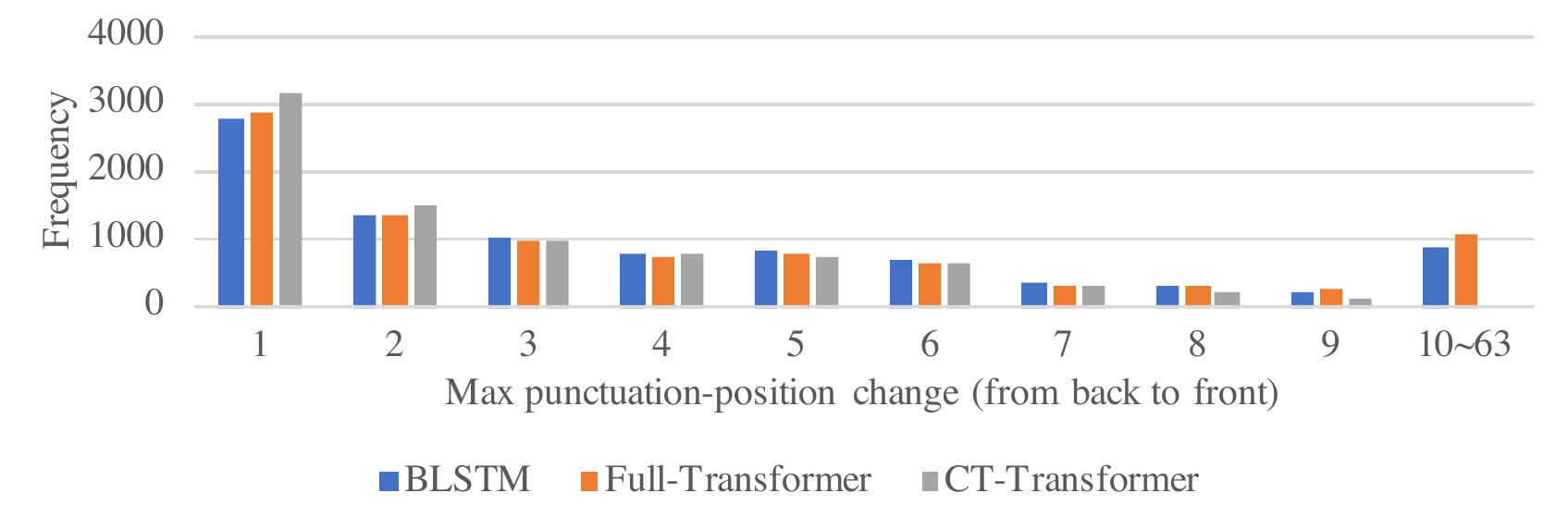}
\caption{The frequency of the max punctuation-position change (from back to front) for BSLTM, Full-Transformer, and CT-Transformer.}
\label{fig:analysis}
\end{figure}

Figure~\ref{fig:analysis} shows the histogram of the max punctuation-position change during decoding on the Chinese test set for the three models. The upper limit of the max position change is 9 for CT-Transformer, but up to 63 for BLSTM and up to 42 for Full-Transformer. BLSTM and Full-Transformer both have about 10\% cases of 10+ max position change. 
These results verify that the proposed CT-Transformer can indeed control time delay which is difficult for Full-Transformer.

\vspace{-3mm}
\section{Conclusions}
We propose Controllable Time-delay Transformer (CT-Transformer) to jointly model punctuation prediction and disfluency detection, which facilitates freezing partial outputs with controllable time-delay to fulfill the real-time constraints in partial decoding required by subsequent applications. We further propose a fast decoding strategy to reduce the latency while maintaining competitive performance, and explore transfer learning to utilize existing well-formed text.
Experimental results demonstrate that CT-Transformer outperforms previous state-of-the-art models on both F-score and latency on the English IWSLT2011 benchmark and an in-house Chinese dataset. Future work includes improving the robustness of our models on ASR transcripts and multilingual transcripts.

\small{
\bibliographystyle{IEEEbib}
\bibliography{refs}
}

\end{document}